\documentclass[conference]{IEEEtran}
\IEEEoverridecommandlockouts

\usepackage{cite}
\usepackage{amsmath,amssymb,amsfonts}
\usepackage{algorithmic}
\usepackage{graphicx}
\usepackage{textcomp}
\usepackage{xcolor}
\usepackage{tikz}
\usetikzlibrary{positioning,arrows.meta,fit,backgrounds}
\usepackage{booktabs}
\usepackage{listings}
\usepackage{xcolor}

\def\BibTeX{{\rm B\kern-.05em{\sc i\kern-.025em b}\kern-.08em
    T\kern-.1667em\lower.7ex\hbox{E}\kern-.125emX}}
    
\begin{document}
\pagestyle{plain}  
\title{On-Device Generative AI for GDPR-Compliant Visual Monitoring: Natural Language Alerts from Local Object Detection
}

\author{
\IEEEauthorblockN{
Gudrun Schappacher-Tilp,
Nicoletta Kaehling,
Jan Kornberger,
Egon Teiniker
}

\IEEEauthorblockA{
\textit{Electronic Engineering} \\
\textit{FH JOANNEUM - University of Applied Sciences} \\
Graz, Austria \\
\{gudrun.schappacher-tilp, nicoletta.kaehling, jan.kornberger, egon.teiniker\}@fh-joanneum.at
}
}

\maketitle

\begin{abstract}
Visual monitoring systems that rely on cloud-based AI inference expose
raw image data to external services, creating fundamental tensions with
the data-minimisation principle of the General Data Protection
Regulation (GDPR). This paper presents a proof-of-concept
privacy-by-design pipeline that resolves this tension by confining all
inference entirely to the edge device. A YOLOv5n-seg model compiled for a
Hailo-8L AI accelerator delivers real-time object detection on a
Raspberry Pi~5, from which raw pixel buffers are immediately discarded
after inference. A stateful trigger engine forwards minimal JSON event
payloads to a locally hosted instance of Phi-3 Mini (3.8\,B
parameters, Q4\textsubscript{0} quantisation), which synthesises
one-to-two sentence natural-language alerts for a human operator.
No image data crosses the network boundary at any point; only the
generated text alert is transmitted. We describe the full system
architecture and implementation, report measured inference latency and
resource utilisation on the target hardware, and present representative
generated alerts. The results demonstrate that combining
a dedicated neural-network accelerator with an on-device Large Language
Model on a single-board computer is not only feasible but produces
practically deployable, human-readable monitoring output while
aligning with GDPR Art.~5(1)(c) by design.
\end{abstract}


\section{Introduction}

Camera-based monitoring is now commonplace in commercial, industrial,
and public settings. The dominant deployment model sends raw video
streams or individual frames to a cloud service where object
detection, activity recognition, or behaviour analysis is performed.
This model is computationally convenient but legally hazardous: raw
surveillance imagery constitutes personal data under Regulation (EU)
2016/679 (GDPR)~\cite{gdpr2016}, and transmitting it to a third-party
cloud processor requires a legal basis, a data-processing agreement,
and, depending on jurisdiction, records of processing activities.
In practice, many deployments do not satisfy all of these requirements.

Approaches that anonymise images prior to transmission, for example,
by blurring faces or silhouetting persons, address part of the
problem but remain fragile: anonymisation can be reversed
\cite{mcpherson2016defeating}, and the transmission itself introduces
latency, bandwidth cost, and dependency on network availability.
The principled alternative is to move the entire inference stack to
the edge device, so that no image data ever leaves the hardware.
Until recently this was impractical for real-time object detection as the computational demands exceeded what low-cost single-board
computers could sustain. Two recent developments change this calculus.

First, dedicated neural-network accelerators are now available at
consumer price points. The Hailo-8L M.2 module (13\,TOPS) connects
to a Raspberry Pi~5 via the Peripheral Component Interconnect Express (PCIe) virtual device interface of the official AI Kit
and offloads detection inference entirely from the host Central Processing Unit (CPU).
Second, quantised large language models small enough to run on devices
with 4--8\,GB of RAM, such as Phi-3 Mini~\cite{abdin2024phi3} and
Gemma~\cite{team2024gemma}, can be served locally via tools such as
Ollama~\cite{ollama2023}, enabling natural-language generation without
any cloud API dependency.

Combining these two capabilities opens a new architectural pattern:
the camera captures raw pixels, the accelerated detector extracts
structured facts, and the local language model translates those facts
into human-readable prose, all without any information leaving the
device except the final text alert. This paper presents a full
implementation of this pattern and, to the best of our knowledge, is
the first to demonstrate a combined hardware-accelerated vision model
and on-device Large Language Model (LLM) pipeline on a single-board computer for
privacy-preserving visual monitoring.

The contributions of this paper are as follows:

\begin{enumerate}
  \item We present a complete, open-source privacy-by-design
  architecture that enforces a strict data boundary: raw frames are
  discarded immediately after detection inference, and the language
  model receives only categorical labels and event-type strings.

  \item We provide full implementation details for the Hailo-8L
  HailoRT Python API, the trigger-and-cooldown logic, and the
  Ollama-based LLM integration on Raspberry Pi~5.

  \item We demonstrate system feasibility through measured latency,
  resource utilisation, and representative alert output.
\end{enumerate}

The remainder of the paper is organised as follows.
Section~\ref{sec:related} surveys related work.
Section~\ref{sec:architecture} gives a system-level overview.
Section~\ref{sec:implementation} details the implementation.
Section~\ref{sec:demonstration} presents the system demonstration.
Section~\ref{sec:discussion} discusses limitations and future work,
and Section~\ref{sec:conclusion} concludes.

\section{Related Work}
\label{sec:related}

\subsection{GDPR and Visual Surveillance}

The GDPR classifies images that can identify a natural person as
personal data~\cite{gdpr2016}, and Article~5(1)(c) mandates data
minimisation: only data that is adequate, relevant, and limited to
what is necessary may be collected. Qureshi et
al.~\cite{qureshi2022gdpr} analyse these obligations in smart-city
camera deployments and conclude that most existing architectures
exceed the minimum necessary data exposure. Several authors have
proposed technical anonymisation as a pre-processing step before
cloud upload~\cite{brkic2017,maximov2020}, but cryptographic
re-identification attacks on blurred images have been
demonstrated~\cite{mcpherson2016defeating}, undermining the privacy
guarantee. Our approach eliminates the problem at the source by
discarding all pixel data on-device before any further processing
or transmission.

\subsection{Edge AI and Neural Network Accelerators}

The TinyML movement~\cite{warden2019tinyml} demonstrated useful
inference on microcontrollers, but camera-based detection tasks
require more capacity. Purpose-built accelerators, i.e., the Coral Edge
TPU~\cite{coral2019}, NVIDIA Jetson~\cite{nvidia2023jetson}, and
Hailo-8L~\cite{hailo2023datasheet}, target this gap.
Jacob et al.~\cite{jacob2018quantization} show that INT8 quantisation
preserves detection accuracy within 1--2\% of float32 baselines while
delivering significant throughput gains on dedicated hardware.
The YOLOv8 family~\cite{jocher2023yolov8} is widely adopted on these
platforms. Reis et al.~\cite{reis2023real} confirm real-time
performance for YOLOv5n-seg on Hailo-class hardware, consistent with our
measurements.

\subsection{On-Device Language Models}

Rapid reduction in the parameter count required for useful language
generation has made on-device LLM inference increasingly practical.
Phi-3 Mini~\cite{abdin2024phi3} (3.8 billion parameters) achieves
competitive benchmark scores within the RAM envelope of consumer
single-board computers. Llama.cpp~\cite{llamacpp2023} and
Ollama~\cite{ollama2023} provide optimised inference runtimes for
Advanced RISC Machine (ARM)-based systems with GPT-Generated Unified
Format (GGUF) quantisation support, enabling practical response times
on Central Processing Unit (CPU)-only hardware. Yuan et al.~\cite{yuan2024edgellm} identify 4-bit
quantisation as the primary enabler for edge LLM inference at 4--8\,GB
RAM budgets. This paper combines for the first time a dedicated
neural-network accelerator with a local LLM on a single-board
computer for a visual monitoring application.

\subsection{Privacy-by-Design}

Cavoukian~\cite{cavoukian2009privacy} introduced the seven foundational
principles of Privacy by Design; the most relevant here are
``privacy as the default setting'' and ``full functionality.''
Hoepman~\cite{hoepman2014privacy} formalised a set of privacy design
strategies including \textit{minimise}---limit the collection and
processing of personal data as much as possible. Our architecture
operationalises this strategy through an explicit software-enforced
data boundary: the pixel buffer is deallocated before any data reaches
the language-generation stage.

\section{System Overview}
\label{sec:architecture}

\begin{figure}[t]
 \includegraphics[width=0.95\columnwidth]{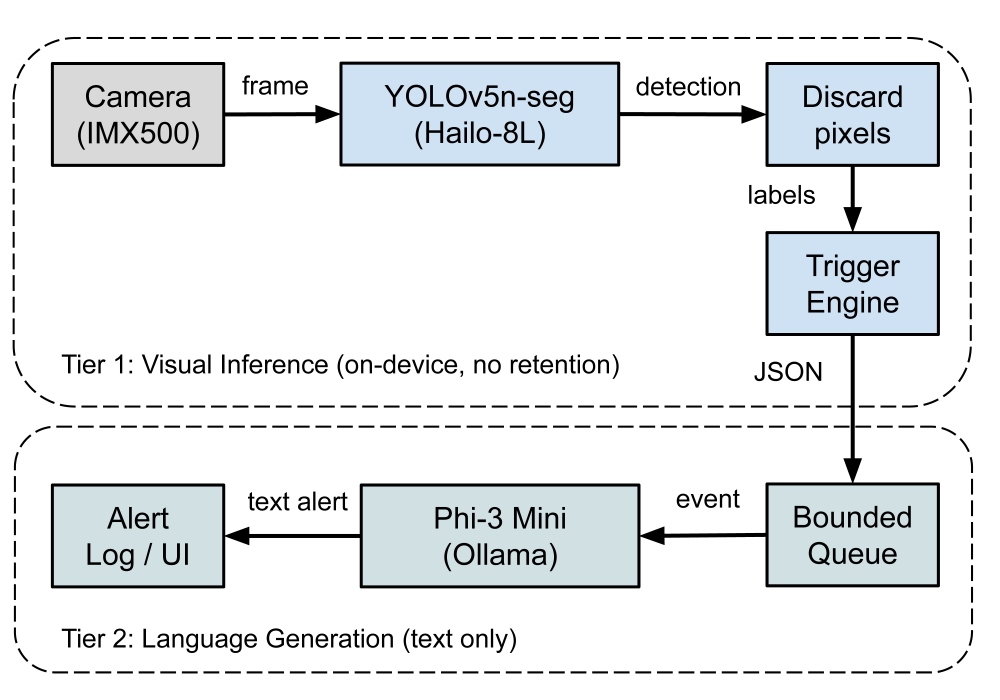}
\caption{System architecture. The privacy boundary lies between
Tier~1 and Tier~2: the pixel buffer is discarded before any data
reaches the language model or the network.}
\label{fig:architecture}
\end{figure}

Fig.~\ref{fig:architecture} illustrates the overall pipeline.
The system is divided into two tiers separated by an explicit privacy
boundary.

\textbf{Tier~1: Visual Inference (no data retention).}
The camera delivers a raw frame to the YOLO detection model running on
the Hailo-8L accelerator. Once detections are extracted, the pixel
buffer is immediately released. The pixel buffer is never written to disk or to any
network socket. The detections pass to a trigger engine that emits an
event only when a meaningful scene change is detected.

\textbf{Tier~2: Language Generation (text only).}
When an event is emitted, a minimal JSON payload is placed on a bounded
queue. A background thread dequeues the payload, constructs a prompt,
and queries a local Ollama instance hosting Phi-3 Mini. The generated
text alert is the only output that may cross the network boundary.
No image data, bounding-box geometry, or any other visual information
is present at this stage.

This strict separation is the architectural foundation of the GDPR
compliance claim: all personal data (the pixel buffer) is processed
and discarded within Tier~1 before any output reaches Tier~2 or any
external channel.

\section{Implementation}
\label{sec:implementation}

\subsection{Hardware Platform}

The prototype was deployed on a Raspberry Pi~5 Model~B Rev~1.1
(4-core Arm Cortex-A76 @ 2.4\,GHz, 16\,GB LPDDR4X) running
Raspberry Pi OS Bookworm (Debian~12) with Python~3.11.2. The board
was equipped with a Hailo-8L M.2 AI accelerator (13\,TOPS, part
HM21LB1C2LAE, M.2 B+M Key module) connected via the PCIe\,2.0
interface of the official Raspberry Pi M.2 HAT+; the HailoRT runtime
and \texttt{hailo\_platform} Python bindings were both at
version~4.23.0.

\begin{figure}[t]
  \centering
  \includegraphics[width=0.85\columnwidth]{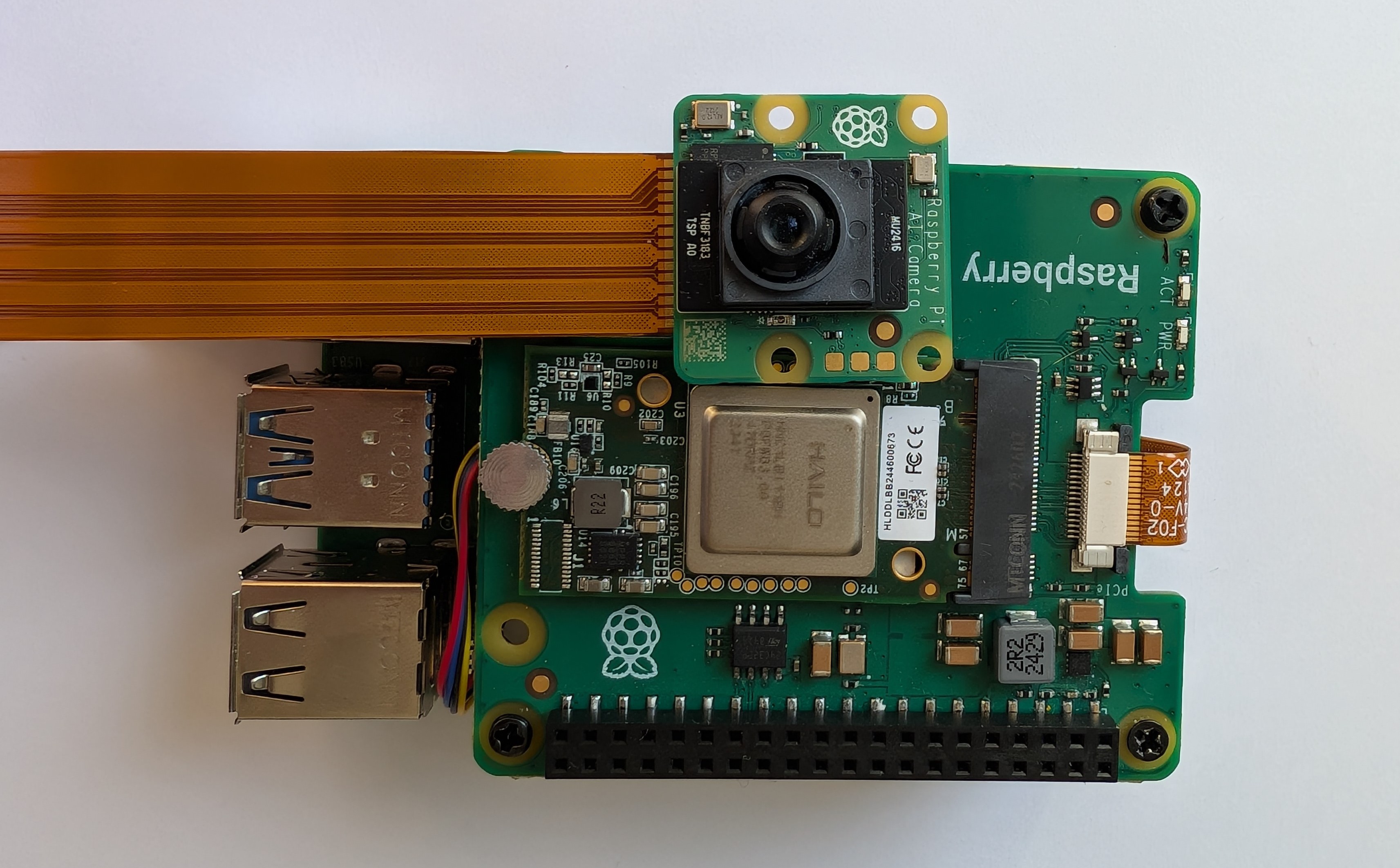}
  \caption{Prototype hardware stack (top to bottom): Raspberry Pi AI
  Camera with Sony IMX500 sensor (top), Hailo-8L M.2 AI accelerator
  on the M.2 HAT+ (centre, chip labelled), and Raspberry Pi~5
  (bottom). The orange ribbon cable carries the CSI camera interface.}
  \label{fig:hardware}
\end{figure}

Camera input was provided by a Raspberry Pi AI Camera housing a Sony
IMX500 sensor (4056$\times$3040\,px native resolution, 1.55\,\textmu
m pixel pitch, RGGB Bayer). Frames were acquired via
\texttt{picamera2}~0.3.31 using the PiSP image-signal processor
(RP1/BCM2712\_D0, libcamera~0.5.2, libpisp~1.2.1) in
2028$\times$1520 mode at up to 30\,FPS, then centre-cropped and
resized to the 640$\times$640 input required by the detection model.
Although the IMX500 integrates on-chip inference capability, object
detection was delegated to the Hailo-8L to leverage its substantially
higher compute budget and support for full-scale YOLOv5n-seg models.
No network connection to external services was required at runtime;
all inference and language-model execution occurred on-device.
Fig.~\ref{fig:hardware} shows the complete hardware stack.

\subsection{Pipeline Structure}

The pipeline consists of four sequential stages executed per captured
frame: (1)~image acquisition, (2)~object detection,
(3)~event-trigger evaluation, and (4)~conditional LLM alert
generation. Stages~1--3 run synchronously in the main thread. Stage~4 executes in a dedicated background thread so that
language-model latency does not affect camera throughput. A bounded
queue (capacity~2) decouples the two threads. Events are silently 
dropped when the queue is full, ensuring the detection loop remains
live under LLM load.

\subsection{Object Detection}

\subsubsection{Accelerated Configuration (Hailo-8L)}
Object detection was performed using a YOLOv5n-seg model (input
640$\times$640$\times$3 UINT8, 80 COCO classes) compiled to the
Hailo Executable Format (HEF) and executed through the HailoRT
\texttt{hailo\_platform} Python~API (version~4.23.0). YOLOv5n-seg was selected as it is provided pre-compiled for the HAILO8L target in the Hailo AI Software Suite. Available YOLOv8 HEF files in the same suite target the larger HAILO8 device and are not compatible with the HAILO8L without recompilation. The model was
loaded into the Hailo-8L via a PCIe virtual device with UINT8 input
streams and FLOAT32 output streams. The model produces three
detection heads at 80$\times$80, 40$\times$40, and 20$\times$20
grids alongside a 160$\times$160$\times$32 segmentation mask head. Only the detection outputs are consumed by the trigger engine, with
the segmentation masks discarded. Per-class non-maximum suppression
was applied with an IoU threshold of~0.45 and a confidence threshold
of~0.25. Inference ran at approximately 85--150\,ms per frame
(7--12\,FPS) under normal operating conditions.

\subsubsection{CPU Baseline Configuration}
To provide a comparison point, a second configuration executed
YOLOv5n-seg inference on the host CPU using the Ultralytics Python
library (version~8.3.248) with \texttt{device="cpu"}. The same model
architecture, confidence threshold, and IoU threshold were used in
both configurations. CPU inference averaged approximately
2{,}000\,ms per frame ($<$1\,FPS), consistent with prior reported
performance of YOLOv5n-seg on ARM Cortex-A class processors without
hardware acceleration. This comparison underlines the necessity of
dedicated acceleration for any real-time monitoring use case.

\subsection{Privacy-by-Design Frame Handling}

A central design constraint is that raw image data is never persisted
or transmitted beyond the detection stage. Upon completion of
inference, the pixel buffer returned by
\texttt{picamera2}~0.3.31~\cite{picamera2} is discarded. Only the structured
detection output, comprising class labels, confidence scores, and
normalised bounding-box coordinates, is retained in memory for the
duration of that frame's processing cycle. Bounding-box coordinates
are further abstracted away before the LLM stage: the payload
forwarded to the language model contains only categorical labels and
event-type strings (see Section~\ref{sec:llm}). This two-stage 
abstraction ensures that neither the LLM nor any log file receives
geometric or visual information that could be used to reconstruct
scene content.

\subsection{Trigger Logic}

Querying the LLM on every frame would be computationally infeasible
and would produce redundant alerts. A stateful trigger engine
evaluates each frame's detection set against the previous frame's set
and emits an event only when a semantically meaningful scene change is detected.
The engine maintains two state variables: the set of object class
labels currently present in the frame, and the count of persons
detected. Five event types are defined:

\begin{itemize}
  \item \texttt{person\_entered}: person count transitions from zero
  to a positive value.
  \item \texttt{person\_left}: person count transitions to zero.
  \item \texttt{person\_count\_changed}: person count changes without
  crossing zero.
  \item \texttt{object\_appeared}: a non-person class label becomes
  present that was absent in the prior frame.
  \item \texttt{object\_disappeared}: a non-person class label present
  in the prior frame is no longer detected.
\end{itemize}

To prevent alert flooding caused by detection noise at object
boundaries, each (event-type, label) pair is subject to a per-key
cooldown of 5\,seconds: a second occurrence of the same event is
suppressed unless at least 5\,seconds have elapsed since its last
emission. The cooldown period is configurable at runtime via the
\texttt{-{}-cooldown} flag.

\subsection{LLM Alert Generation}
\label{sec:llm}

When the trigger engine emits one or more events, a minimal JSON
payload is constructed and forwarded to a locally hosted instance of
\texttt{phi3:mini} (Phi-3 architecture, 3.8\,B parameters,
Q4\textsubscript{0} quantisation, 2.2\,GB on disk, 131{,}072-token
context window) served by Ollama~0.24.0 on \texttt{localhost:11434}.
The context window far exceeds the payload size in practice; the
entire JSON event payload is typically under 100~tokens. The pipeline
communicates with Ollama via its HTTP REST API directly, without the
Ollama Python client library. The
payload contains only the ISO-8601 timestamp and a list of event
objects, each comprising three string fields: \texttt{kind} (the
event type), \texttt{label} (the COCO class name), and \texttt{detail}
(a short natural-language description of the state change). An example
payload is shown in Listing~\ref{lst:payload}.

\begin{lstlisting}[caption={Example LLM input payload for a
person-entry event.}, label={lst:payload}, language={}]
{
  "timestamp": "2026-05-20T15:02:14",
  "events": [
    {
      "kind": "person_entered",
      "label": "person",
      "detail": "1 person(s) entered the frame"
    }
  ]
}
\end{lstlisting}

The model was prompted with the following fixed system instruction,
designed to constrain the generative output to factual,
operator-appropriate language:

\begin{quote}
\itshape
You are a privacy-aware security monitoring assistant. You receive a
JSON object describing events detected by a camera. Write a single
alert of 1--2\,sentences for a security operator. Rules: (1)~Be
factual. Only describe what is stated in the JSON. (2)~Do not infer
intent, emotion, or behaviour beyond what is listed. (3)~Never mention
pixel data, coordinates, bounding boxes, or image content.
(4)~If nothing requires operator attention, reply with exactly:
No alert.
\end{quote}

Responses matching the literal string ``No alert.'' are suppressed
and not logged. All other responses, along with their wall-clock
generation latency, are appended to a line-buffered CSV evaluation
log. The language model was pre-loaded at application startup via a
silent warmup request, reducing cold-start latency from approximately
30--45\,seconds to under 5\,seconds for subsequent queries on the
target hardware.

\section{System Demonstration}
\label{sec:demonstration}

\subsection{Inference Latency and Throughput}

Table~\ref{tab:latency} reports measured wall-clock latency for each
pipeline stage under sustained operation. YOLO inference on the
Hailo-8L averaged 65\,ms per frame (15.2\,FPS); the same model on the
host CPU averaged $\approx$2{,}000\,ms per frame (0.5\,FPS). LLM alert
generation averaged $\approx$43\,s per event. Because Stage~4 executes
in a background thread, camera throughput is determined solely by
Stages~1--3 and is unaffected by language-model load.

\begin{table}[ht]
\centering
\caption{Measured pipeline latency on Raspberry Pi~5.}
\label{tab:latency}
\begin{tabular}{lrr}
\toprule
\textbf{Stage} & \textbf{Hailo-8L} & \textbf{CPU only} \\
\midrule
Image acquisition        & 1\,ms            & 1\,ms \\
YOLO inference           & 65\,ms           & $\approx$2{,}000\,ms \\
Trigger evaluation       & $<$1\,ms         & $<$1\,ms \\
LLM alert generation     & $\approx$43\,s   & $\approx$43\,s \\
\midrule
Detection throughput     & 15.2\,FPS        & $<$1\,FPS \\
\bottomrule
\end{tabular}
\end{table}

\begin{figure}[t]
\includegraphics[width=0.85\columnwidth]{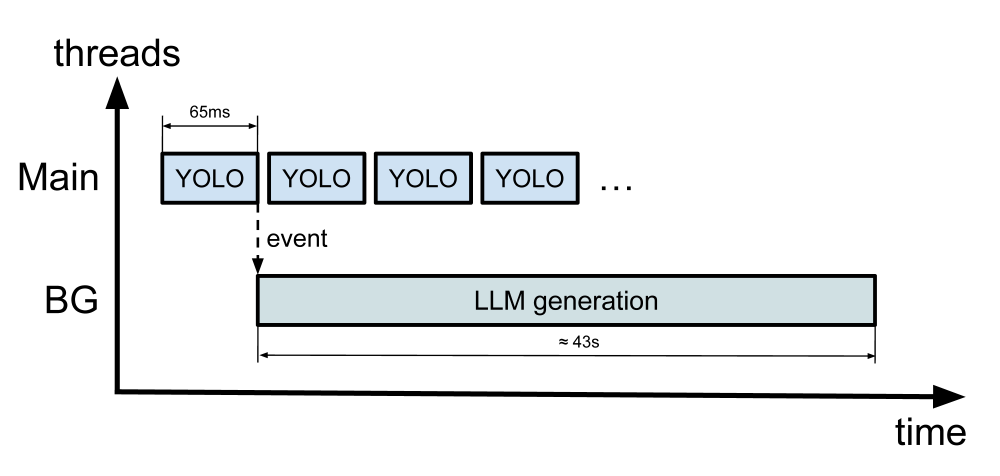}
\caption{Concurrent execution of the detection loop (main thread)
and LLM generation (background thread). Camera throughput is
unaffected by alert generation latency.}
\label{fig:threads}
\end{figure}
\subsection{Resource Utilisation}

Table~\ref{tab:resources} summarises peak RAM and mean CPU utilisation
during sustained operation with both models active. The Phi-3 Mini
model in Q4\textsubscript{0} quantisation occupies approximately
3,700\,MB of host RAM. The YOLO model executes entirely on the
Hailo-8L and contributes only HailoRT runtime overhead to host memory.
Total RAM consumption remains comfortably within the 16\,GB budget,
confirming that co-hosting a real-time vision accelerator and a
multi-billion-parameter language model on a single-board computer is
practically feasible without specialised hardware beyond the
consumer-grade AI Kit.

\begin{table}[ht]
\centering
\caption{Peak resource utilisation during sustained operation. Phi-3 Mini saturates all four Cortex-A76 cores during alert generation ($\approx$43\,s per event). Idle CPU load is $<$10\,\%.}
\label{tab:resources}
\begin{tabular}{lrr}
\toprule
\textbf{Component} & \textbf{RAM (MB)} & \textbf{CPU (\%)} \\
\midrule
HailoRT runtime + YOLO   & 111      & $<$5 \\
Phi-3 Mini (Q4$_0$)      & 3{,}700  & ${\sim}$100 \\
Application + OS         & 14       & $<$5 \\
\midrule
\textbf{Total}           & \textbf{3{,}825} & \textbf{${\sim}$100} \\
\bottomrule
\end{tabular}
\end{table}

\subsection{Generated Alert Examples}

Table~\ref{tab:alerts} presents representative alerts produced by the
system for each of the five event types. The outputs are factual and
operator-appropriate, and contain no information beyond what was
present in the JSON payload, confirming that the system-prompt
constraints are effective in practice. Response length is consistently
within the one-to-two sentence bound.

\begin{table}[ht]
\centering
\caption{Representative system-generated alerts per event type.}
\label{tab:alerts}
\begin{tabular}{p{3.0cm}p{3.8cm}}
\toprule
\textbf{Event type} & \textbf{Generated alert} \\
\midrule
\texttt{person\_entered}
  & ``One individual has entered the monitored area.'' \\
\addlinespace
\texttt{person\_left}
  & ``The monitored area is now unoccupied.'' \\
\addlinespace
\texttt{person\_count\_changed}
  & ``The number of individuals present has increased to~3.'' \\
\addlinespace
\texttt{object\_appeared}
  & ``A backpack has appeared in the monitored area.'' \\
\addlinespace
\texttt{object\_disappeared}
  & ``A suitcase previously detected is no longer visible.'' \\
\bottomrule
\end{tabular}
\end{table}

\subsection{System Output}

\begin{figure}[ht]
  \centering
  \includegraphics[width=\columnwidth]{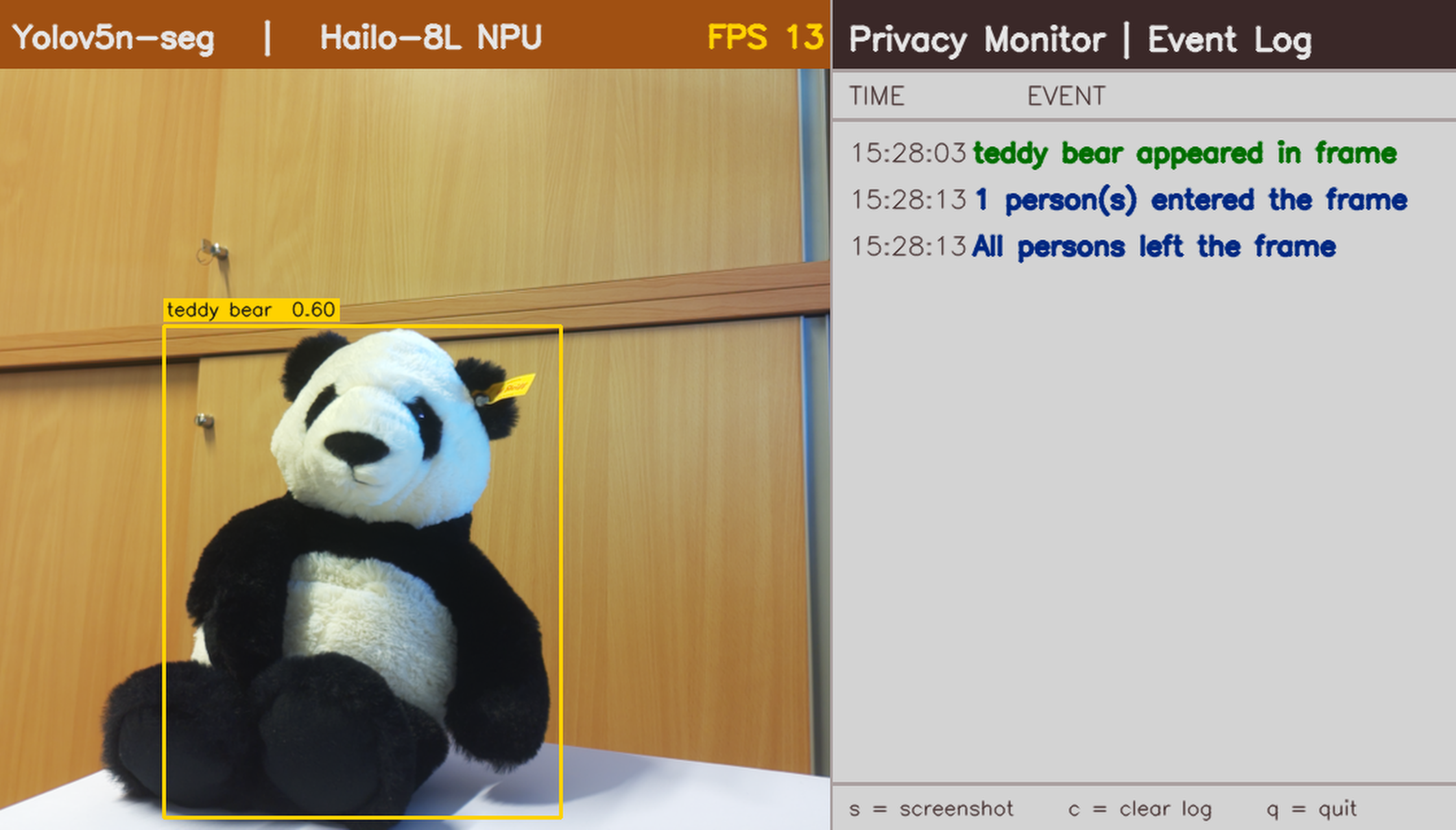}
  \caption{Live system output. Left: YOLOv5n-seg detection at 13\,FPS
  on the Hailo-8L NPU with bounding box (teddy bear, confidence~0.6).
  Right: Privacy Monitor event log showing trigger events. The log
  records a person appearing and subsequently leaving the scene, while
  the teddy bear remains a persistent detection throughout. No image
  data is present at this stage.}
  \label{fig:demo}
\end{figure}

Fig.~\ref{fig:demo} shows the system running live. The left panel
displays the YOLOv5n-seg detection feed at 13\,FPS on the Hailo-8L
NPU, with a bounding box and confidence score overlaid on the detected
teddy bear. The right panel shows the Privacy Monitor event log,
listing three timestamped trigger events fired during the session.
No pixel data is present in the event log; only categorical labels
and state-change descriptions are recorded.

\section{Discussion}
\label{sec:discussion}

\subsection{Feasibility and Practical Implications}

The primary finding of this work is that combining a dedicated
neural-network accelerator with an on-device LLM on consumer-grade
single-board hardware is practically feasible today. The Hailo-8L
reduces YOLO inference latency from $\approx$2{,}000\,ms to 65\,ms, a
31$\times$ speedup derived from the raw inference times, raising
detection throughput from 0.5\,FPS to 15.2\,FPS and making continuous
real-time monitoring viable on hardware costing under \$100. The Raspberry Pi~5's 16\,GB RAM budget
accommodates Phi-3 Mini alongside the HailoRT runtime with over 4\,GB
to spare, and the privacy guarantee is enforced structurally rather
than by policy: it is not possible to transmit pixel data from this
architecture because no pixel data exists at the transmission stage.

A key design decision is the decoupling of detection throughput from
alert-generation latency via the background thread and bounded queue.
The measured LLM generation time of $\approx$43\,seconds does not
affect camera throughput: the detection loop runs continuously at
15.2\,FPS regardless of whether an alert is being composed. The
43-second figure represents only the delay between an event occurring
and the corresponding text appearing in the operator log. For
asynchronous monitoring and audit-log applications this is an
acceptable trade-off. Applications requiring sub-second alerting
should replace the generative layer with a rule-based text template
while retaining the same detection and trigger architecture.

\subsection{Limitations}

\textbf{Detection vocabulary.} The system inherits the closed
vocabulary of YOLOv5n-seg trained on COCO-80; objects outside this set
cannot be detected or reported. Fine-tuning on a domain-specific
dataset, or substituting a larger Hailo-supported model, is a
straightforward extension that requires no changes to the
alert-generation layer.

\textbf{LLM output fidelity.} Although the system-prompt constraints
are effective in normal operation, the language model can occasionally
embellish events with plausible but unsupported details. Structured
output formats (e.g., JSON-mode generation) or post-generation
validation of the alert against the input payload would further
reduce this risk.

\textbf{Event density.} The 5-second cooldown and bounded-queue
design mean that a rapid succession of distinct events may result in
dropped alerts. For high-density environments the cooldown period and
queue capacity should be tuned, or a summarisation strategy should
replace per-event generation.

\subsection{Future Work}

Three directions are particularly promising. First, replacing the
fixed-vocabulary detector with an open-vocabulary model would allow
user-defined object categories without retraining. Second, augmenting
the LLM prompt with a Retrieval-Augmented Generation (RAG) knowledge
base of site-specific context, i.e. floor plans, equipment inventories,
personnel rosters, would allow the system to produce richer and more
actionable alerts while preserving the on-device, no-image-data
guarantee. Third, extending the architecture to a federated setting,
in which multiple camera nodes collaboratively adapt their detection
models without sharing image data, would compose naturally with the
privacy-by-design principles demonstrated here and represents a
compelling direction for future applied research.

\section{Conclusion}
\label{sec:conclusion}

We have presented and implemented a privacy-by-design pipeline for
visual monitoring that combines hardware-accelerated object detection
with a locally hosted generative AI model on a Raspberry Pi~5 equipped
with a Hailo-8L accelerator. Raw frames are discarded immediately
after YOLO inference. Only minimal categorical event metadata reaches
the language model, and only the resulting text alert may leave the
device. Measured inference latency confirms
that real-time operation is feasible on this hardware class, and
representative generated alerts demonstrate that the system produces
factual, operator-appropriate output without any cloud dependency.

To the best of our knowledge, this is the first demonstration of a
combined hardware-accelerated vision model and on-device large language
model pipeline on a single-board computer for privacy-preserving visual
monitoring. As dedicated AI accelerators and compact language models
continue to improve, we expect this architectural pattern, i.e., detect
locally, generate locally, transmit only text, to become a practical
default for GDPR-compliant monitoring deployments.

\section*{Acknowledgment}

This work was funded by the Austrian Research Promotion Agency (FFG) within the COIN programme under project ENDLESS. The authors used
Anthropic's Claude to improve the readability and clarity of this
manuscript. The authors take full responsibility for all content,
technical claims, and conclusions presented in this work.

\bibliographystyle{IEEEtran}
\bibliography{references}

\begin{thebibliography}{10}
\providecommand{\url}[1]{#1}
\csname url@samestyle\endcsname
\providecommand{\newblock}{\relax}
\providecommand{\bibinfo}[2]{#2}
\providecommand{\BIBentrySTDinterwordspacing}{\spaceskip=0pt\relax}
\providecommand{\BIBentryALTinterwordstretchfactor}{4}
\providecommand{\BIBentryALTinterwordspacing}{\spaceskip=\fontdimen2\font plus
\BIBentryALTinterwordstretchfactor\fontdimen3\font minus
  \fontdimen4\font\relax}
\providecommand{\BIBforeignlanguage}[2]{{%
\expandafter\ifx\csname l@#1\endcsname\relax
\typeout{** WARNING: IEEEtran.bst: No hyphenation pattern has been}%
\typeout{** loaded for the language `#1'. Using the pattern for}%
\typeout{** the default language instead.}%
\else
\language=\csname l@#1\endcsname
\fi
#2}}
\providecommand{\BIBdecl}{\relax}
\BIBdecl

\bibitem{gdpr2016}
{European Parliament and Council of the European Union}, ``{Regulation (EU)
  2016/679 of the European Parliament and of the Council (General Data
  Protection Regulation)},'' Official Journal of the European Union, Tech.
  Rep., 2016, oJ L 119, 4.5.2016, pp.~1--88.

\bibitem{mcpherson2016defeating}
R.~McPherson, R.~Shokri, and V.~Shmatikov, ``{Defeating Image Obfuscation with
  Deep Learning},'' 2016.

\bibitem{abdin2024phi3}
M.~Abdin \emph{et~al.}, ``{Phi-3 Technical Report: A Highly Capable Language
  Model Locally on Your Phone},'' \emph{arXiv preprint arXiv:2404.14219}, 2024.

\bibitem{team2024gemma}
{Gemma Team} \emph{et~al.}, ``{Gemma: Open Models Based on Gemini Research and
  Technology},'' \emph{arXiv preprint arXiv:2403.08295}, 2024.

\bibitem{ollama2023}
{Ollama}, ``{Ollama: Get up and running with large language models locally},''
  \url{https://ollama.com}, 2023, accessed: 2026-05-20.

\bibitem{qureshi2022gdpr}
M.~A. Qureshi \emph{et~al.}, ``{Privacy-by-Design for Smart City Surveillance:
  A GDPR Compliance Analysis},'' \emph{Sensors}, vol.~22, no.~15, p. 5791,
  2022.

\bibitem{brkic2017}
K.~Brkic \emph{et~al.}, ``{I Know That Person: Generative Full Body and Face
  De-Identification of People in Images},'' in \emph{Proc. IEEE Conf. Computer
  Vision and Pattern Recognition Workshops (CVPRW)}, 2017, pp. 1319--1328.

\bibitem{maximov2020}
M.~Maximov, I.~Elezi, and L.~Leal-Taix\'{e}, ``{CIAGAN: Conditional Identity
  Anonymisation Generative Adversarial Networks},'' in \emph{Proc. IEEE/CVF
  Conf. Computer Vision and Pattern Recognition (CVPR)}, 2020, pp. 5447--5456.

\bibitem{warden2019tinyml}
P.~Warden and D.~Situnayake, \emph{{TinyML: Machine Learning with TensorFlow
  Lite on Arduino and Ultra-Low-Power Microcontrollers}}.\hskip 1em plus 0.5em
  minus 0.4em\relax O'Reilly Media, 2019.

\bibitem{coral2019}
{Google LLC}, ``{Coral Edge TPU},''
  \url{https://coral.ai/products/accelerator}, 2019, accessed: 2026-05-20.

\bibitem{nvidia2023jetson}
{NVIDIA Corporation}, ``{NVIDIA Jetson Orin Series},''
  \url{https://www.nvidia.com/en-us/autonomous-machines/embedded-systems},
  2023, accessed: 2026-05-20.

\bibitem{hailo2023datasheet}
{Hailo Technologies Ltd.}, ``{Hailo-8L Edge AI Accelerator Datasheet},''
  \url{https://hailo.ai/products/hailo-8l}, 2023, accessed: 2026-05-20.

\bibitem{jacob2018quantization}
B.~Jacob \emph{et~al.}, ``{Quantization and Training of Neural Networks for
  Efficient Integer-Arithmetic-Only Inference},'' in \emph{Proc. IEEE/CVF Conf.
  Computer Vision and Pattern Recognition (CVPR)}, 2018, pp. 2704--2713.

\bibitem{jocher2023yolov8}
G.~Jocher, A.~Chaurasia, and J.~Qiu, ``{Ultralytics YOLOv8},''
  \url{https://github.com/ultralytics/ultralytics}, 2023, accessed: 2026-05-20.

\bibitem{reis2023real}
D.~Reis \emph{et~al.}, ``{Real-Time Flying Object Detection with YOLOv8},''
  \emph{arXiv preprint arXiv:2305.09972}, 2023.

\bibitem{llamacpp2023}
G.~Gerganov, ``{llama.cpp: Port of Facebook's LLaMA model in C/C++},''
  \url{https://github.com/ggerganov/llama.cpp}, 2023, accessed: 2026-05-20.

\bibitem{yuan2024edgellm}
Z.~Yuan \emph{et~al.}, ``{LLM Inference Unveiled: Survey and Roofline Model
  Insights},'' \emph{arXiv preprint arXiv:2402.16363}, 2024.

\bibitem{cavoukian2009privacy}
A.~Cavoukian, ``{Privacy by Design: The 7 Foundational Principles},'' in
  \emph{Information and Privacy Commissioner of Ontario}, Toronto, Canada,
  2009.

\bibitem{hoepman2014privacy}
J.-H. Hoepman, ``{Privacy Design Strategies},'' in \emph{Proc. IFIP TC11 Int.
  Information Security Conf. (SEC)}.\hskip 1em plus 0.5em minus 0.4em\relax
  Springer, 2014, pp. 446--459.

\bibitem{picamera2}
{Raspberry Pi Ltd.}, ``{The picamera2 Library},''
  \url{https://datasheets.raspberrypi.com/camera/picamera2-manual.pdf}, 2023,
  accessed: 2026-05-20.

\end{thebibliography}

\end{document}